\documentclass{article}

\PassOptionsToPackage{numbers, sort&compress}{natbib}
\usepackage[final]{nips_2017}

\usepackage[utf8]{inputenc} 
\usepackage[T1]{fontenc}    
\usepackage{hyperref}       
\usepackage{url}            
\usepackage{booktabs}       
\usepackage{amsfonts}       
\usepackage{nicefrac}       
\usepackage{microtype}      
\usepackage{bm}
\usepackage[mathscr]{eucal}
\usepackage{epstopdf}
\usepackage{makecell}
\usepackage{multirow}
\usepackage{graphicx}
\usepackage[ruled,linesnumbered]{algorithm2e}
\usepackage{wrapfig}
\usepackage{subcaption}
\usepackage{javen}

\title{Meta-SGD: Learning to Learn Quickly\\ for Few-Shot Learning}

\author{
  Zhenguo Li \hspace{0.8cm} Fengwei Zhou \hspace{0.8cm} Fei Chen \hspace{0.8cm} Hang Li 
    \\ 
  Huawei Noah's Ark Lab\\
  \texttt{\{li.zhenguo, zhou.fengwei, chenfei100, hangli.hl\}@huawei.com} \\
}

\begin{document}

\maketitle

\begin{abstract}
Few-shot learning is challenging for learning algorithms that learn each task in isolation and from scratch. In contrast, meta-learning learns from many related tasks a meta-learner that can learn a new task more accurately and faster with fewer examples, where the choice of meta-learners is crucial.
In this paper, we develop Meta-SGD, an SGD-like, easily trainable meta-learner that can initialize and adapt any differentiable learner in just one step, on both supervised learning and reinforcement learning. Compared to the popular meta-learner LSTM, Meta-SGD is conceptually simpler, easier to implement, and can be learned more efficiently. Compared to the latest meta-learner MAML, Meta-SGD has a much higher capacity by learning to learn not just the learner initialization, but also the learner update direction and learning rate, all in a single meta-learning process.
Meta-SGD shows highly competitive performance for few-shot learning on regression, classification, and reinforcement learning.
\end{abstract}

\section{Introduction}

The ability to learn and adapt rapidly from small data is essential to intelligence. However,
current success of deep learning relies greatly on big labeled data.
It learns each task in isolation and from scratch, by fitting a deep neural network over data
through extensive, incremental model updates using stochastic gradient descent (SGD). The approach is inherently data-hungry and time-consuming, with fundamental challenges for problems with limited data or in dynamic environments where fast adaptation is critical.
In contrast, humans can learn quickly from a few examples
by leveraging prior experience. Such capacity in data efficiency and fast adaptation, if realized in machine learning, can greatly expand its utility. This motivates the study of few-shot learning, which aims to learn quickly from only a few examples~\cite{lake2016building}.

Several existing ideas may be adapted for few-shot learning. In transfer learning, one often fine-tunes a pre-trained model using target data~\cite{sharif2014cnn}, where it is challenging not to unlearn the previously acquired knowledge. In multi-task learning, the target task is trained jointly with auxiliary ones to distill inductive bias about the target problem~\cite{caruana1998multitask}. It is tricky to decide what to share in the joint model. In semi-supervised learning, one augments labeled target data with massive unlabeled data to leverage a holistic distribution of the data~\cite{wu2012learning}. Strong assumptions are required for this method to work. While these efforts can alleviate the issue of data scarcity to some extend, the way prior knowledge is used is specific and not generalizable. A principled approach for few-shot learning to representing, extracting and leveraging prior knowledge is in need.

Meta-learning offers a new perspective to machine learning, by lifting the learning level from data to tasks ~\cite{schmidhuber1987evolutionary,bengio1991learning,thrun2012learning}. Consider supervised learning. The common practice learns from a set of labeled examples, while meta-learning learns from a set of (labeled) tasks, each represented as a labeled training set and a labeled testing set. The hypothesis is that by being exposed to a broad scope of a task space, a learning agent may figure out a learning strategy tailored to the tasks in that space.

Specifically, in meta-learning, a learner for a specific task is learned by a learning algorithm called meta-learner, which is learned on a bunch of similar tasks to maximize the combined generalization power of the learners of all tasks. The learning occurs at two levels and in different time-scales. Gradual learning is performed across tasks, which learns a meta-learner to carry out rapid learning within each task, whose feedback is used to adjust the learning strategy of the meta-learner. Interestingly, the learning process can continue forever, thus enabling life-long learning, and at any moment, the meta-learner can be applied to learn a learner for any new task. Such a two-tiered learning to learn strategy for meta-learning has been applied successfully to few-shot learning on  classification~\cite{santoro2016meta,vinyals2016matching,ravi2017optimization,finn2017model}, regression~\cite{santoro2016meta,finn2017model}, and reinforcement learning~\cite{wang2016learning,Yan2016reinf,finn2017model,mishra2017meta,sung2017learning}.

The key in meta-learning is in the design of meta-learners to be learned. In general terms,
a meta-learner is a trainable learning algorithm that can train a learner,
influence its behavior, or itself function as a learner.
Meta-learners developed so far include recurrent models~\cite{hochreiter2001learning,santoro2016meta,wang2016learning,Yan2016reinf}, metrics~\cite{koch2015siamese,vinyals2016matching}, or optimizers~\cite{andrychowicz2016learning,li2016learning,ravi2017optimization,finn2017model}.
A recurrent model such as Long Short-Term Memory (LSTM)~\cite{hochreiter1997long} processes data sequentially and figures out its own learning strategy from scratch in the course~\cite{santoro2016meta}. Such meta-learners are versatile but less comprehensible, with applications in classification~\cite{santoro2016meta}, regression~\cite{hochreiter2001learning,santoro2016meta}, and reinforcement learning~\cite{wang2016learning,Yan2016reinf}.
A metric influences a learner by modifying distances between examples. Such meta-learners are more suitable for non-parametric learners such as the $k$-nearest neighbors algorithm or its variants~\cite{vinyals2016matching,koch2015siamese}. Meta-learners above do not learn an explicit learner, which is typically done by an optimizer such as SGD. This suggests that optimizers, if trainable, can serve as meta-learners. The meta-learner perspective of optimizers, which is used to be hand-designed, opens the door for learning optimizers via meta-learning.

\begin{figure}
  \centering
  \includegraphics[width=0.48\textwidth]{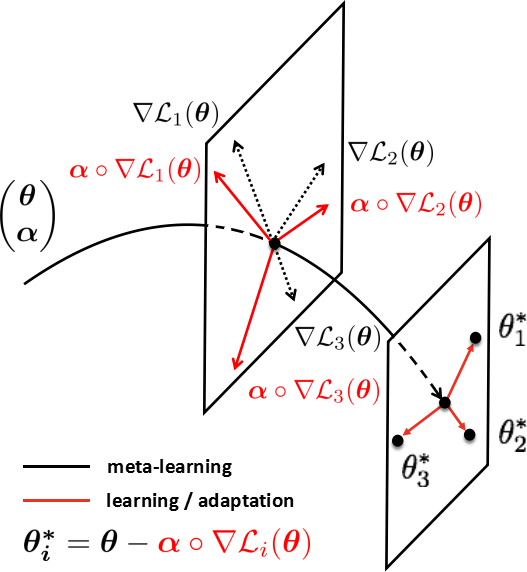}
  \caption{Illustrating the two-level learning process of Meta-SGD. Gradual learning is performed across tasks at the meta-space $(\thetab, \alphab)$ that learns the meta-learner. Rapid learning is carried out by the meta-learner in the learner space $\thetab$ that learns task-specific learners.
  }
  \label{fig:meta-learning-process}
  \vspace{-0pt}
\end{figure}

Recently, LSTM is used to update models such as Convolutional Neural Network (CNN) iteratively like SGD~\cite{andrychowicz2016learning,ravi2017optimization}, where both initialization and update strategy are learned via meta-learning, thus called Meta-LSTM in what follows. This should be in sharp contrast to SGD where the initialization is randomly chosen, the learning rate is set manually, and the update direction simply follows the gradient. While Meta-LSTM shows promising results on few-shot learning~\cite{ravi2017optimization} or as a generic optimizer~\cite{andrychowicz2016learning}, it is rather difficult to train. In practice, each parameter of the learner is updated independently in each step, which greatly limits its potential. In this paper, we develop a new optimizer that is very easy to train. Our proposed  meta-learner acts like SGD, thus called Meta-SGD (Figure~\ref{fig:meta-learning-process}), but the initialization, update direction, and learning rates are learned via meta-learning, like Meta-LSTM. Besides much easier to train than Meta-LSTM, Meta-SGD also learns much faster than Meta-LSTM. It can learn effectively from a few examples even in one step. Experimental results on regression, classification, and reinforcement learning unanimously show that Meta-SGD is highly competitive on few-show learning.

\section{Related Work}

One popular approach to few-shot learning is with generative models, where one notable work is by~\cite{lake2015human}. It uses probabilistic programs to represent concepts of handwritten characters, and exploits the specific knowledge of how pen strokes are composed to produce characters. This work shows how knowledge of related concepts can ease learning of new concepts from even one example, using the principles of compositionality and learning to learn~\cite{lake2016building}.

A more general approach to few-shot learning is by meta-learning, which trains a meta-learner from many related tasks to direct the learning of a learner for a new task,
without relying on ad hoc knowledge about the problem.
The key is in developing high-capacity yet trainable meta-learners.
\cite{vinyals2016matching} suggest metrics as meta-learners for non-parametric learners such as $k$-nearest neighbor classifiers. Importantly, it matches training and testing conditions in meta-learning, which works well for few-shot learning and is widely adopted afterwards. Note that a metric does not really train a learner, but influences its behavior by modifying distances between examples. As such, metric meta-learners mainly work for non-parametric learners.

Early studies show that a recurrent neural network (RNN) can model adaptive optimization algorithms~\cite{cotter1990fixed,younger1999fixed}. This suggests its potential as meta-learners. Interestingly, \cite{hochreiter2001learning} find that LSTM performs best as meta-learner among various architectures of RNNs. \cite{andrychowicz2016learning} formulate LSTM as a generic, SGD-like optimizer which shows promising results compared to widely used hand-designed optimization algorithms. In \cite{andrychowicz2016learning}, LSTM is used to imitate the model update process of the learner (e.g., CNN) and output model increment at each timestep.  \cite{ravi2017optimization} extend \cite{andrychowicz2016learning} for few-shot learning, where the
LSTM cell state represents the parameters of the learner and the variation of the cell state corresponds to model update (like gradient descent) of the learner. Both initialization and update strategy are learned jointly \cite{ravi2017optimization}. However, using LSTM as meta-learner to learn a learner such as CNN incurs prohibitively high complexity. In practice, each parameter of the learner is updated independently in each step, which may significantly limit its potential. \cite{santoro2016meta} adapt a memory-augmented LSTM~\cite{graves2014neural} for few-shot learning, where the learning strategy is figured out as the LSTM rolls out. \cite{finn2017model} use SGD as meta-learner, but only the initialization is learned. Despite its simplicity, it works well in practice.

\section{Meta-SGD}


\subsection{Meta-Learner}


In this section, we propose a new meta-learner that applies to both supervised learning (i.e., classification and regression) and reinforcement learning. For simplicity, we use supervised learning as running case and discuss reinforcement learning later. How can a meta-learner $M_{\phib}$ initialize and adapt a learner $f_{\thetab}$ for a new task from a few examples $\mathcal{T}=\{(\xb_i,\yb_i)\}$?
One standard way updates the learner iteratively from random initialization using gradient descent:
\begin{equation}\label{euq:typical_updates}
\thetab^{t}  = \thetab^{t-1} - \alpha \nabla\mathcal{L}_{\mathcal{T}}(\thetab^{t-1}),
\end{equation}
where $\mathcal{L}_{\mathcal{T}}(\thetab)$ is the empirical loss
$$
\mathcal{L}_{\mathcal{T}}(\thetab)= \frac{1}{|\mathcal{T}|}\sum_{(\xb,\yb)\in \mathcal{T}} \ell(f_{\thetab}(\xb), \yb)
$$
with some loss function $\ell$, $\nabla\mathcal{L}_{\mathcal{T}}(\thetab)$ is the gradient of $\mathcal{L}_{\mathcal{T}}(\thetab)$, and $\alpha$ denotes the learning rate that is often set manually.

With only a few examples, it is non-trivial to decide how to initialize and when to stop the learning process to avoid overfitting. Besides, while gradient is an effective direction for data fitting, it may lead to overfitting under the few-shot regime. This also makes it tricky to choose the learning rate. While many ideas may be applied for regularization, it remains challenging to balance between the induced prior and the few-shot fitting. What in need is a principled approach that determines all learning factors in a way that maximizes generalization power rather than data fitting. Another important aspect regards the speed of learning: can we learn within a couple of iterations? Besides an interesting topic on its own~\cite{lake2015human}, this will enable many emerging applications such as self-driving cars and autonomous robots that require to learn and react in a fast changing environment.

The idea of learning to learn appears to be promising for few-shot learning. Instead of hand-designing a learning algorithm for the task of interest, it learns from many related tasks how to learn, which may include how to initialize and update a learner, among others, by training a meta-learner to do the learning. The key here is in developing a high-capacity yet trainable meta-learner. While other meta-learners are possible, here we consider meta-learners in the form of optimizers, given their broad generality and huge success in machine learning. Specifically, we aim to learn an optimizer for few-shot learning.

There are three key ingredients in defining an optimizer: initialization, update direction, and learning rate. The initialization is often set randomly, the update direction often follows gradient or some variant (e.g., conjugate gradient), and the learning rate is usually set to be small, or decayed over iterations. While such rules of thumb work well with a huge amount of labeled data, they are unlikely reliable for few-shot learning. In this paper, we present a meta-learning approach that automatically determines all the ingredients of an optimizer in an end-to-end manner.

Mathematically, we propose the following meta-learner composed of an initialization term and an adaptation term:
\begin{align}\label{eq:meta-SGD}
\bm{\theta}^\prime = \bm{\theta} - \bm{\alpha}\circ\nabla\mathcal{L}_{\mathcal{T}}(\bm{\theta}),
\end{align}
where $\thetab$ and $\alphab$ are (meta-)parameters of the meta-learner to be learned, and $\circ$ denotes element-wise product. Specifically, $\thetab$ represents the state of a learner that can be used to initialize the learner for any new task, and $\alphab$ is a vector of the same size as $\thetab$ that decides both the update direction and learning rate. The adaptation term $\bm{\alpha}\circ\nabla\mathcal{L}_{\mathcal{T}}(\bm{\theta})$ is a vector whose direction represents the update direction and whose length represents the learning rate. Since the direction of $\bm{\alpha}\circ\nabla\mathcal{L}_{\mathcal{T}}(\bm{\theta})$ is usually different from that of the gradient $\nabla\mathcal{L}_{\mathcal{T}}(\bm{\theta})$, it implies that the meta-learner does not follow the gradient direction to update the learner, as does by SGD. Interestingly, given $\alphab$, the adaptation is indeed fully determined by the gradient, like SGD.

In summary, given a few examples $\mathcal{T}=\{(\xb_i,\yb_i)\}$ for a few-shot learning problem, our meta-learner first initializes the learner with $\thetab$ and then adapts it to $\thetab^\prime$ in just one step, in a new direction $\bm{\alpha}\circ\nabla\mathcal{L}_{\mathcal{T}}(\bm{\theta})$ different from the gradient $\nabla\mathcal{L}_{\mathcal{T}}(\bm{\theta})$ and using a learning rate implicitly implemented in $\bm{\alpha}\circ\nabla\mathcal{L}_{\mathcal{T}}(\bm{\theta})$. As our meta-learner also relies on the gradient as in SGD but it is learned via meta-learning rather than being hand-designed like SGD, we call it Meta-SGD.


\subsection{Meta-training}

\begin{figure}[t]
	\centering
	\includegraphics[width=\textwidth]{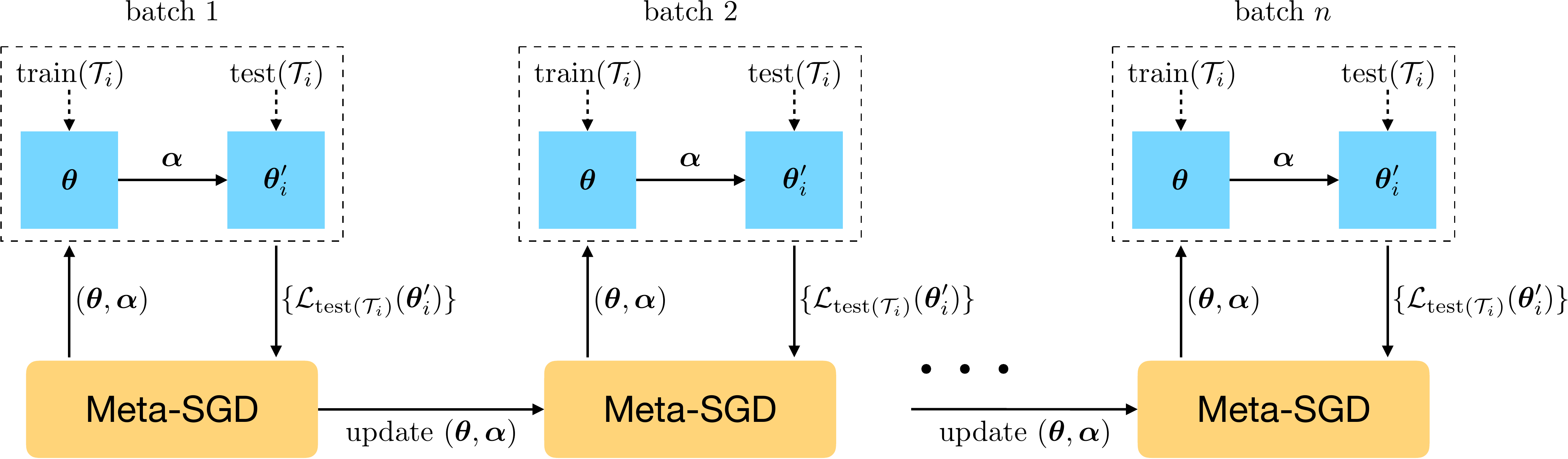}\vspace{0.0cm}
	\caption{Meta-training process of Meta-SGD.
	}
	\label{fig:meta-update}\vspace{-10pt}
\end{figure}

We aim to train the meta-learner to perform well on many related tasks. For this purpose, assume there is a distribution $p(\mathcal{T})$ over the related task space, from which we can randomly sample tasks. A task $\mathcal{T}$ consists of a training set $\mathrm{train}(\mathcal{T})$ and a testing set $\mathrm{test}(\mathcal{T})$.
Our objective is to maximize the expected generalization power of the meta-learner on the task space. Specifically, given a task $\mathcal{T}$ sampled from $p(\mathcal{T})$, the meta-learner learns the learner based on the training set $\mathrm{train}(\mathcal{T})$, but the generalization loss is measured on the testing set $\mathrm{test}(\mathcal{T})$. Our goal is to train the meta-learner to minimize the expected generalization loss.

Mathematically, the learning of our meta-learner is formulated as the optimization problem as follows:
\begin{align}
\min_{\bm{\theta},\bm{\alpha}} E_{\mathcal{T}\sim p(\mathcal{T})}[\mathcal{L}_{\mathrm{test}(\mathcal{T})}(\bm{\theta}^\prime)]
=E_{\mathcal{T}\sim p(\mathcal{T})}[\mathcal{L}_{\mathrm{test}(\mathcal{T})}(\bm{\theta}-\bm{\alpha}\circ \nabla\mathcal{L}_{\mathrm{train}(\mathcal{T})}(\bm{\theta}))].
\end{align}

The above objective is differentiable w.r.t. both $\thetab$ and $\alphab$, which allows to use SGD to solve it efficiently, as shown in Algorithm~\ref{algo:train} and illustrated in Figure~\ref{fig:meta-update}.



\IncMargin{1.5em}
\begin{algorithm}[h]
	\caption{\small Meta-SGD for Supervised Learning}
	\label{algo:train}
	\SetNlSty{textbf}{}{:}
	\SetAlgoNoLine
	
	\Indm
	\KwIn{  task distribution $p(\CMcal{T})$, learning rate $\beta$
	}
	\KwOut{$\ \thetab, \alphab$
	}

	\Indp
	Initialize $\thetab, \alphab$\;
    \While {not done}{
    Sample batch of tasks $\CMcal{T}_i\sim p(\CMcal{T})$\;
	\For{all $\mathcal{T}_i$}{
        \mbox{$\mathcal{L}_{\mathrm{train}(\mathcal{T}_i)}(\thetab)\leftarrow \frac{1}{|\mathrm{train}(\CMcal{T}_i)|}\sum\limits_{(\xb, \yb) \in \mathrm{train}(\CMcal{T}_i)} \ell(f_{\thetab}(\xb), \yb);$}\\
        $\thetab_{i}^\prime \leftarrow \thetab - \alphab \circ \nabla \mathcal{L}_{\mathrm{train}(\mathcal{T}_i)}(\thetab);$\\
        \mbox{$\mathcal{L}_{\mathrm{test}(\mathcal{T}_i)}(\thetab_i^\prime) \leftarrow
                \frac{1}{|\mathrm{test}(\CMcal{T}_i)|}\sum\limits_{(\xb,
                  \yb)\in \mathrm{test}(\CMcal{T}_i)} \ell(f_{\thetab_{i}^\prime}(\xb), \yb);$}
	}
$(\thetab, \alphab) \leftarrow (\thetab, \alphab) - \beta \nabla_{(\thetab, \alphab)} \sum_{\CMcal{T}_i} \mathcal{L}_{\mathrm{test}(\mathcal{T}_i)}(\thetab_i^\prime);$
}
	
\end{algorithm}

\textbf{Reinforcement Learning.}
In reinforcement learning, we regard a task as a Markov decision process (MDP). Hence, a task $\mathcal{T}$ contains a tuple $(\mathcal{S}, \mathcal{A}, q, q_0, T, r, \gamma)$, where $\mathcal{S}$ is a set of states, $\mathcal{A}$ is a set of actions, $q:\mathcal{S}\times\mathcal{A}\times\mathcal{S}\rightarrow [0,1]$ is the transition probability distribution, $q_0:\mathcal{S}\rightarrow [0,1]$ is the initial state distribution, $T\in\mathbb{N}$ is the horizon, $r:\mathcal{S}\times\mathcal{A}\rightarrow\mathbb{R}$ is the reward function, and $\gamma\in [0,1]$ is the discount factor. The learner $f_{\bm{\theta}}:\mathcal{S}\times\mathcal{A}\rightarrow [0,1]$ is a stochastic policy, and the loss $\mathcal{L}_{\mathcal{T}}(\bm{\theta})$ is the negative expected discounted reward
\begin{align}\label{eq:reward}
\mathcal{L}_{\mathcal{T}}(\bm{\theta})=-\mathbb{E}_{\bm{s_t},\bm{a_t} \sim f_{\bm{\theta}},q,q_0}\left[\sum_{t=0}^T \gamma^t r(\bm{s_t},\bm{a_t})\right].
\end{align}
As in supervised learning, we train the meta-learner to minimize the expected generalization loss. Specifically, given a task $\mathcal{T}$ sampled from $p(\mathcal{T})$, we first sample $N_1$ trajectories according to the policy $f_{\bm{\theta}}$. Next, we use policy gradient methods to compute the empirical policy gradient $\nabla\mathcal{L}_{\mathcal{T}}(\bm{\theta})$ and then apply equation~\ref{eq:meta-SGD} to get the updated policy $f_{\bm{\theta}^\prime}$. After that, we sample $N_2$ trajectories according to $f_{\bm{\theta}^\prime}$ and compute the generalization loss.

The optimization problem for reinforcement learning can be rewritten as follows:
\begin{align}
\min_{\bm{\theta},\bm{\alpha}} E_{\mathcal{T}\sim p(\mathcal{T})}[\mathcal{L}_{\mathcal{T}}(\bm{\theta}^\prime)]
=E_{\mathcal{T}\sim p(\mathcal{T})}[\mathcal{L}_{\mathcal{T}}(\bm{\theta}-\bm{\alpha}\circ \nabla\mathcal{L}_{\mathcal{T}}(\bm{\theta}))],
\end{align}
and the algorithm is summarized in Algorithm~\ref{algo:train_reinforcement}.

\IncMargin{1.5em}
\begin{algorithm}[h]
	\caption{\small \mbox{Meta-SGD for Reinforcement Learning}}
	\label{algo:train_reinforcement}
	\SetNlSty{textbf}{}{:}
	\SetAlgoNoLine
	
	\Indm
	\KwIn{  task distribution $p(\CMcal{T})$, learning rate $\beta$
	}
	\KwOut{$\ \thetab, \alphab$
	}

	\Indp
	Initialize $\thetab, \alphab$\;
    \While {not done}{
    Sample batch of tasks $\CMcal{T}_i\sim p(\CMcal{T})$\;
	\For{all $\mathcal{T}_i$}{
        Sample $N_1$ trajectories according to $f_{\bm{\theta}}$;\\
        Compute policy gradient $\nabla\mathcal{L}_{\mathcal{T}_i}(\bm{\theta})$;\\
        $\bm{\theta}_{i}^\prime \leftarrow \bm{\theta} - \bm{\alpha} \circ \nabla \mathcal{L}_{\mathcal{T}_i}(\bm{\theta})$;\\
        Sample $N_2$ trajectories according to $f_{\bm{\theta}_i^\prime}$;\\
        \mbox{Compute policy gradient $\nabla_{(\bm{\theta}, \bm{\alpha})}\mathcal{L}_{\mathcal{T}_i}(\bm{\theta}_i^\prime)$;}\\
	}
$(\thetab, \alphab) \leftarrow (\thetab, \alphab) - \beta \nabla_{(\bm{\theta}, \bm{\alpha})} \sum_{\mathcal{T}_i} \mathcal{L}_{\mathcal{T}_i}(\bm{\theta}_i^\prime);$
}
	
\end{algorithm}

\subsection{Related Meta-Learners}

Let us compare Meta-SGD with other meta-learners in the form of optimizer. MAML~\cite{finn2017model} uses the original SGD as meta-learner, but the initialization is learned via meta-learning. In contrast, Meta-SGD also learns the update direction and the learning rate, and may have a higher capacity. Meta-LSTM~\cite{ravi2017optimization} relies on LSTM to learn all initialization, update direction, and learning rate, like Meta-SGD, but it incurs a much higher complexity than Meta-SGD. In practice, it learns each parameter of the learner independently at each step, which may limit its potential.



\section{Experimental Results}
\label{sec:experiment}

We evaluate the proposed meta-learner Meta-SGD on a variety of few-shot learning problems on regression, classification, and reinforcement learning. We also compare its performance with the state-of-the-art results reported in previous work. Our results show that Meta-SGD can learn very quickly from a few examples with only one-step adaptation.
All experiments are run on Tensorflow~\cite{abadi2016tensorflow}.


\subsection{Regression}

In this experiment, we evaluate Meta-SGD on the problem of $K$-shot regression, and compare it with the state-of-the-art meta-learner MAML~\cite{finn2017model}. The target function is a sine curve $y(x)=A\sin(\omega x + b)$, where the amplitude $A$, frequency $\omega$, and phase $b$ follow the uniform distribution on intervals $[0.1, 5.0]$, $[0.8, 1.2]$, and $[0, \pi]$, respectively. The input range is restricted to the interval $[-5.0, 5.0]$. The $K$-shot regression task is to estimate the underlying sine curve from only $K$ examples.

For meta-training, each task consists of $K\in\{5, 10, 20\}$ training examples and 10 testing examples with inputs randomly chosen from $[-5.0, 5.0]$. The prediction loss is measured by the mean squared error (MSE). For the regressor, we follow \cite{finn2017model} to use a small neural network with an input layer of size 1, followed by 2 hidden layers of size 40 with ReLU nonlinearities, and then an output layer of size 1. All weight matrices use truncated normal initialization with mean 0 and standard deviation 0.01, and all bias vectors are initialized by 0. For Meta-SGD, all entries in $\alphab$ have the same initial value randomly chosen from $[0.005, 0.1]$. For MAML, a fixed learning rate $\alpha=0.01$ is used following \cite{finn2017model}. Both meta-learners use one-step adaptation and are trained for 60000 iterations with meta batch-size of 4 tasks.

For performance evaluation (meta-testing), we randomly sample 100 sine curves. For each curve, we sample $K$ examples for training with inputs randomly chosen from $[-5.0, 5.0]$, and another 100 examples for testing with inputs evenly distributed on $[-5.0, 5.0]$.
We repeat this procedure 100 times and take the average of MSE. The results averaged over the sampled 100 sine curves with $95\%$ confidence intervals are summarized in Table~\ref{tab:res_regression_relu}.

By Table~\ref{tab:res_regression_relu}, Meta-SGD performs consistently better than MAML on all cases with a wide margin, showing that Meta-SGD does have a higher capacity than MAML by learning all the initialization, update direction, and learning rate simultaneously, rather than just the initialization as in MAML. By learning all ingredients of an optimizer across many related tasks, Meta-SGD well captures the problem structure and is able to learn a learner with very few examples.
In contrast, MAML regards the learning rate $\alpha$ as a hyper-parameter and just follows the gradient of empirical loss to learn the learner, which may greatly limit its capacity. Indeed, if we change the learning rate $\alpha$ from 0.01 to 0.1, and re-train MAML via 5-shot meta-training, the prediction losses for 5-shot, 10-shot, and 20-shot meta-testing increase to $1.77 \pm 0.30$, $1.37 \pm 0.23$, and $1.15 \pm 0.20$, respectively.

\begin{figure}[t]
	\centering
	\begin{subfigure}{0.49\textwidth}
		\centering
		\includegraphics[width=1.0\textwidth]{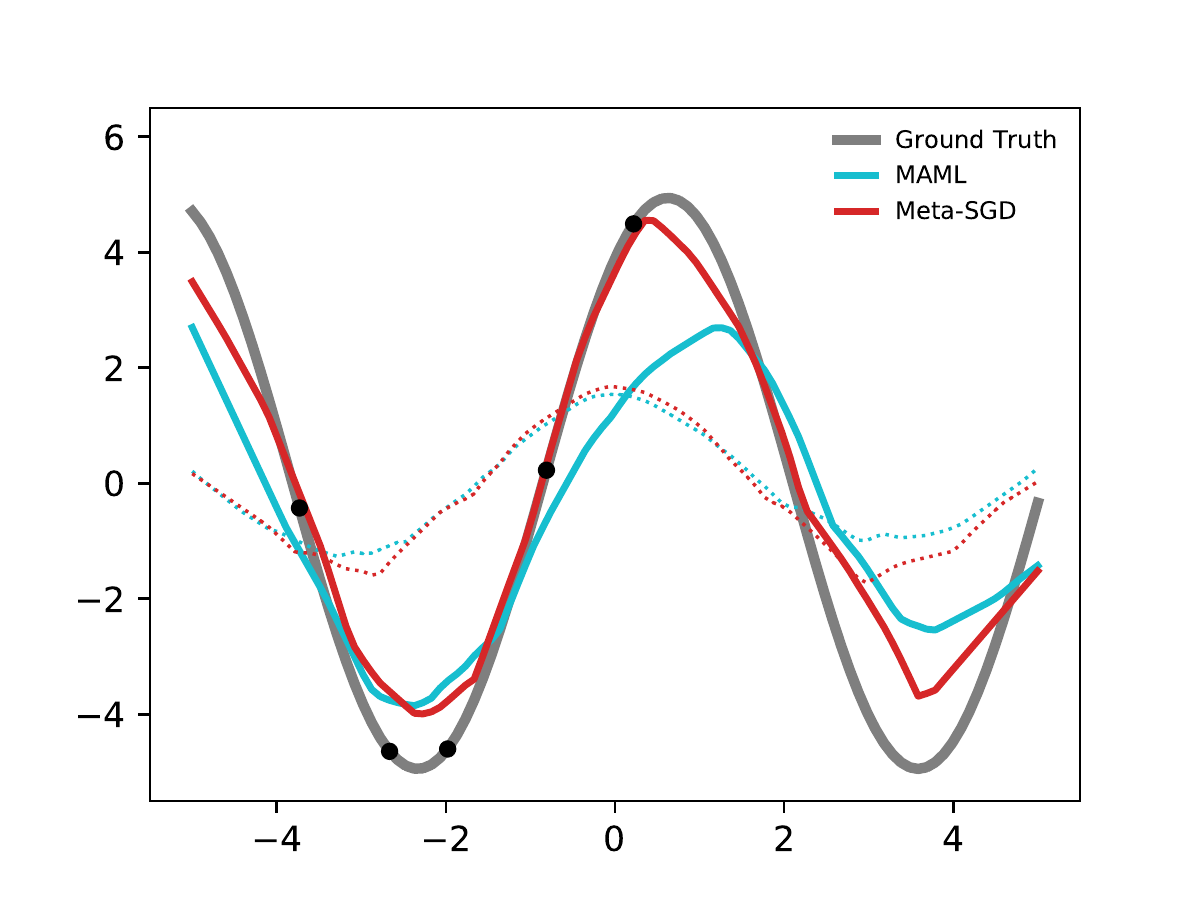}
	\end{subfigure}
	\begin{subfigure}{0.49\textwidth}
		\centering
		\includegraphics[width=1.0\textwidth]{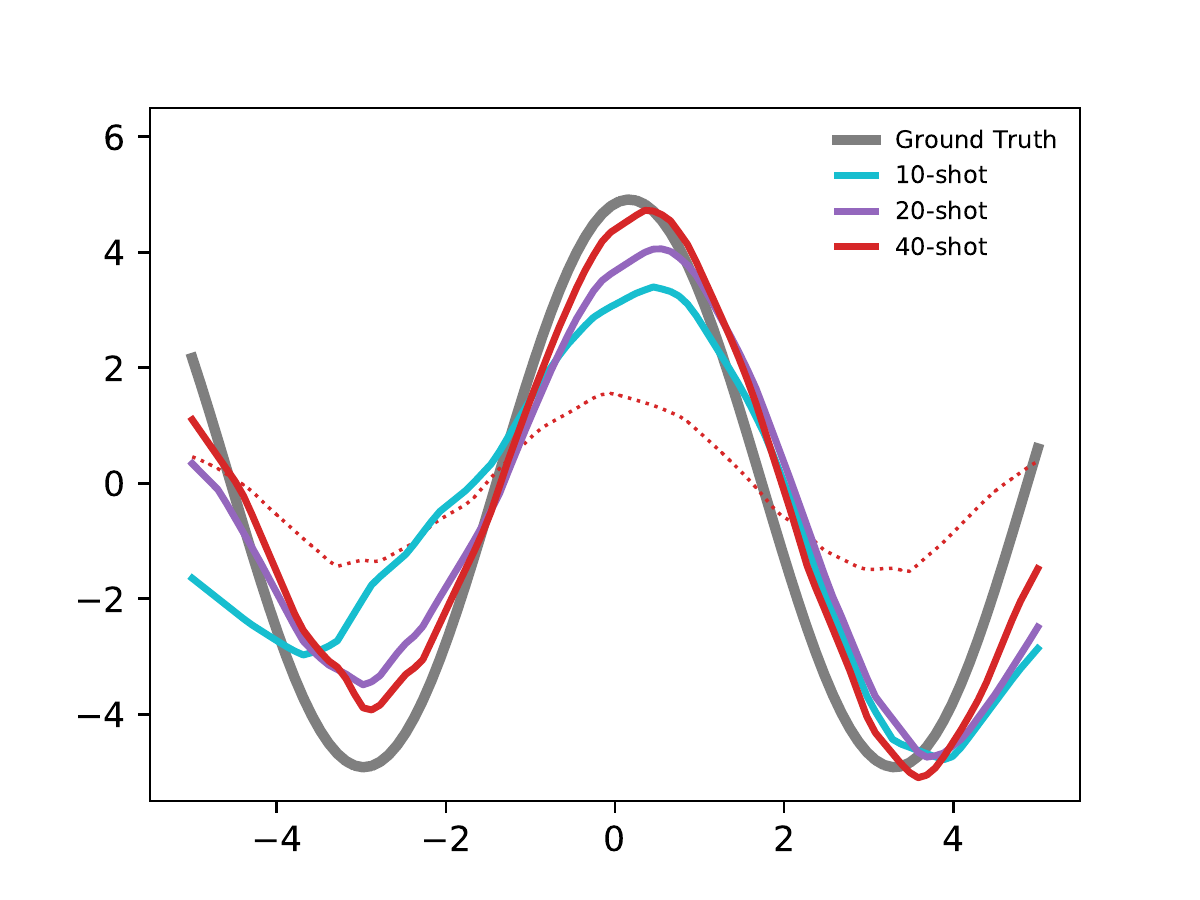}
	\end{subfigure}
	
	\caption{\textbf{Left:} Meta-SGD vs MAML on 5-shot regression. Both initialization (dotted) and result after one-step adaptation (solid) are shown. \textbf{Right:} Meta-SGD (10-shot meta-training) performs better with more training examples in meta-testing. }
	\label{fig:reg_curve}\vspace{-0.0cm}
\end{figure}

Figure~\ref{fig:reg_curve} shows how the meta-learners perform on a random 5-shot regression task. From Figure~\ref{fig:reg_curve} (left), compared to MAML, Meta-SGD can adapt more quickly to the shape of the sine curve after just one step update with only 5 examples, even when these examples are all in one half of the input range. This shows that Meta-SGD well captures the meta-level information across all tasks. Moreover, it continues to improve with additional training examples during meta-tesing, as shown in Figure~\ref{fig:reg_curve} (right). While the performance of MAML also gets better with more training examples, the regression results of Meta-SGD are still better than those of MAML (Table~\ref{tab:res_regression_relu}). This shows that our learned optimization strategy is better than gradient descent even when applied to solve the tasks with large training data.

\begin{table}[t]
	\centering
	\begin{center}
		\caption{Meta-SGD vs MAML on few-shot regression}
				\vspace{0.1cm}
		\label{tab:res_regression_relu}
		\begin{tabular}{|c|c|c|c|c|}
			\hline
			Meta-training & Models & 5-shot testing & 10-shot testing & 20-shot testing \\
			\hline
			\multirow{2}{*}{5-shot training} & MAML & $1.13 \pm 0.18$ & $0.85 \pm 0.14$ & $0.71 \pm 0.12$ \\
			& Meta-SGD & $\mathbf{0.90 \pm 0.16}$ & $\mathbf{0.63 \pm 0.12}$ & $\mathbf{0.50 \pm 0.10}$ \\
			\hline
			\multirow{2}{*}{10-shot training} & MAML & $1.17 \pm 0.16$ & $0.77 \pm 0.11$ & $0.56 \pm 0.08$ \\
			& Meta-SGD & $\mathbf{0.88 \pm 0.14}$ & $\mathbf{0.53 \pm 0.09}$ & $\mathbf{0.35 \pm 0.06}$ \\
			\hline
			\multirow{2}{*}{20-shot training} & MAML & $1.29 \pm 0.20$ & $0.76 \pm 0.12$ & $0.48 \pm 0.08$ \\
			& Meta-SGD & $\mathbf{1.01 \pm 0.17}$ & $\mathbf{0.54 \pm 0.08}$ & $\mathbf{0.31 \pm 0.05}$ \\
			\hline
		\end{tabular}
	\end{center}
	\vspace{-0.5cm}
\end{table}

%



\subsection{Classification}

We evaluate Meta-SGD on few-shot classification using two benchmark datasets Omniglot and MiniImagenet.

\noindent\textbf{Omniglot.} The Omniglot dataset~\cite{lake2011one} consists of 1623 characters from 50 alphabets. Each character contains 20 instances drawn by different individuals. We randomly select 1200 characters for meta-training, and use the remaining characters for meta-testing. We consider 5-way and 20-way classification for both 1 shot and 5 shots.

\noindent\textbf{MiniImagenet.} The MiniImagenet dataset consists of 60000 color images from 100 classes, each with 600 images. The data is divided into three disjoint subsets: 64 classes for meta-training,
16 classes for meta-validation, and 20 classes for meta-testing \cite{ravi2017optimization}. We consider 5-way and 20-way classification for both 1 shot and 5 shots.

We train the model following~\cite{vinyals2016matching}. For an $N$-way $K$-shot classification task, we first sample $N$ classes from the meta-training dataset, and then in each class sample $K$ images for training and $15$ other images for testing. We update the meta-learner once for each batch of tasks. After meta-training, we test our model with unseen classes from the meta-testing dataset.
Following~\cite{finn2017model}, we use a convolution architecture with 4 modules, where each module consists of $3\times 3$ convolutions, followed by batch
normalization~\cite{ioffe2015batch}, a ReLU nonlinearity, and $2\times
2$ max-pooling.
For Omniglot, the images are downsampled to $28 \times 28$, and we use 64 filters and add an additional fully-connected layer with dimensionality 32 after the convolution modules.
For MiniImagenet, the images are downsampled to $84 \times 84$, and we use 32 filters in the convolution modules.

We train and evaluate Meta-SGD that adapts the learner in one step.
In each iteration of meta-training, Meta-SGD is updated once with one batch of tasks.
We follow~\cite{finn2017model} for batch size settings. For Omniglot, the batch size is set to 32 and 16 for 5-way and 20-way classification, respectively. For MiniImagenet, the batch size is set to 4 and 2 for 1-shot and 5-shot classification, respectively. We add a regularization term to the objective function. 

\begin{table}[t]
	\centering
	\begin{center}
		\caption{Classification accuracies on Omniglot}
		\label{tab:res_omniglot}
		\vspace{0.1cm}
		\begin{tabular}{|l|c|c|c|c|}
			\hline
			\multirow{2}{*}{} & \multicolumn{2}{c|}{5-way Accuracy} & \multicolumn{2}{c|}{20-way Accuracy} \\ \cline{2-5}
			
			& 1-shot & 5-shot & 1-shot & 5-shot \\ \hline
			
			Siamese Nets & $97.3\%$ & $98.4\%$ & $88.2\%$ & $97.0\%$ \\ \hline
			
			Matching Nets & $98.1\%$ & $98.9\%$ & $93.8\%$ & $98.5\%$ \\ \hline
			
			MAML & $98.7 \pm 0.4 \% $ & $99.9 \pm 0.1 \% $ & $95.8 \pm 0.3 \% $ & $98.9 \pm 0.2 \% $ \\ \hline
			
			Meta-SGD & $\mathbf{99.53 \pm 0.26 \%} $ & $\mathbf{99.93 \pm 0.09 \%} $ & $\mathbf{95.93 \pm 0.38 \%} $ & $\mathbf{98.97 \pm 0.19 \%} $ \\ \hline
			
		\end{tabular}
	\end{center}
	\vspace{-5pt}
\end{table}

\begin{table}[t]
	\centering
	\begin{center}
		\caption{Classification accuracies on MiniImagenet}
		\label{tab:res_miniimagenet}
		\vspace{0.1cm}
		\begin{tabular}{|l|c|c|c|c|}
			\hline
			\multirow{2}{*}{} & \multicolumn{2}{c|}{5-way Accuracy} & \multicolumn{2}{c|}{20-way Accuracy} \\ \cline{2-5}
			
			& 1-shot & 5-shot & 1-shot & 5-shot \\ \hline
			
			Matching Nets & $43.56 \pm 0.84 \% $ & $55.31 \pm 0.73 \% $ & $17.31 \pm 0.22 \% $ & $22.69 \pm 0.20 \% $ \\ \hline
			
			Meta-LSTM & $43.44 \pm 0.77 \% $ & $60.60 \pm 0.71 \% $ & $16.70 \pm 0.23 \% $ & ${26.06 \pm 0.25 \%} $ \\ \hline
			
			MAML & $48.70 \pm 1.84 \% $ & $63.11 \pm 0.92 \% $ & $16.49 \pm 0.58 \% $ & $19.29 \pm 0.29 \% $ \\ \hline
			
			Meta-SGD & $\mathbf{50.47 \pm 1.87 \%} $ & $\mathbf{64.03 \pm 0.94 \%} $ & $\mathbf{17.56 \pm 0.64 \%}$ & $\mathbf{28.92 \pm 0.35 \%}$ \\ \hline
			
		\end{tabular}
	\end{center}
	\vspace{-5pt}
\end{table}


The results of Meta-SGD are summarized in Table~\ref{tab:res_omniglot} and Table~\ref{tab:res_miniimagenet}, together with results of other state-of-the-art models, including Siamese Nets~\cite{koch2015siamese}, Matching Nets~\cite{vinyals2016matching}, Meta-LSTM~\cite{ravi2017optimization}, and MAML~\cite{finn2017model}. The results of previous models for 5-way and 20-way classification on Omniglot, and 5-way classification on MiniImagenet are reported in previous work~\cite{finn2017model}, while those for 20-way classification on MiniImagenet are obtained in our experiment. For the 20-way results on MiniImagenet, we run Matching Nets and Meta-LSTM using the implementation by~\cite{ravi2017optimization}, and MAML using our own implementation\footnote{The code provided by~\cite{finn2017model} does not scale for this 5-shot 20-way problem in one GPU with 12G memory used in our experiment.}.
For MAML, the learning rate $\alpha$ is set to 0.01 as in the 5-way case, and the learner is updated with one gradient step for both meta-training and meta-testing tasks like Meta-SGD. All models are trained for 60000 iterations.
The results represent mean accuracies with $95\%$ confidence intervals over tasks.

For Omniglot, our model Meta-SGD is slightly better than the state-of-the-art models on all classification tasks.
In our experiments we noted that for 5-shot classification tasks, the model performs better when it is trained with 1-shot tasks during meta-training than trained with 5-shot tasks. This phenomenon was observed in both 5-way and 20-way classification. The 5-shot (meta-testing) results of Meta-SGD in Table~\ref{tab:res_omniglot} are obtained via 1-shot meta-training.


For MiniImagenet, Meta-SGD outperforms all other models in all cases. Note that Meta-SGD learns the learner in just one step, making it faster to train the model and to adapt to new tasks, while still improving accuracies. In comparison, previous models often update the learner using SGD with multiple gradient steps or using LSTM with multiple iterations.
For 20-way classification, the results of Matching Nets shown in Table~\ref{tab:res_miniimagenet} are obtained when the model is trained with 10-way classification tasks. When trained with 20-way classification tasks, its accuracies drop to $12.27 \pm 0.18$ and $21.30 \pm 0.21$ for 1-shot and 5-shot, respectively, suggesting that Matching Nets may need more iterations for sufficient training, especially for 1-shot.
We also note that for 20-way classification, MAML with the learner updated in one gradient step performs worse than Matching Nets and Meta-LSTM. In comparison, Meta-SGD has the highest accuracies for both 1-shot and 5-shot.
We also run experiments on MAML for 5-way classification where the learner is updated with 1 gradient step for both meta-training and meta-testing, the mean accuracies of which are $44.40\%$ and $61.11\%$ for 1-shot and 5-shot classification, respectively.
These results show the capacity of Meta-SGD in terms of learning speed and performance for few-shot classification.

\subsection{Reinforcement Learning}

\begin{figure*}[t]
	\centering
	\begin{subfigure}{0.49\textwidth}
		\centering
		\includegraphics[width=1.0\textwidth]{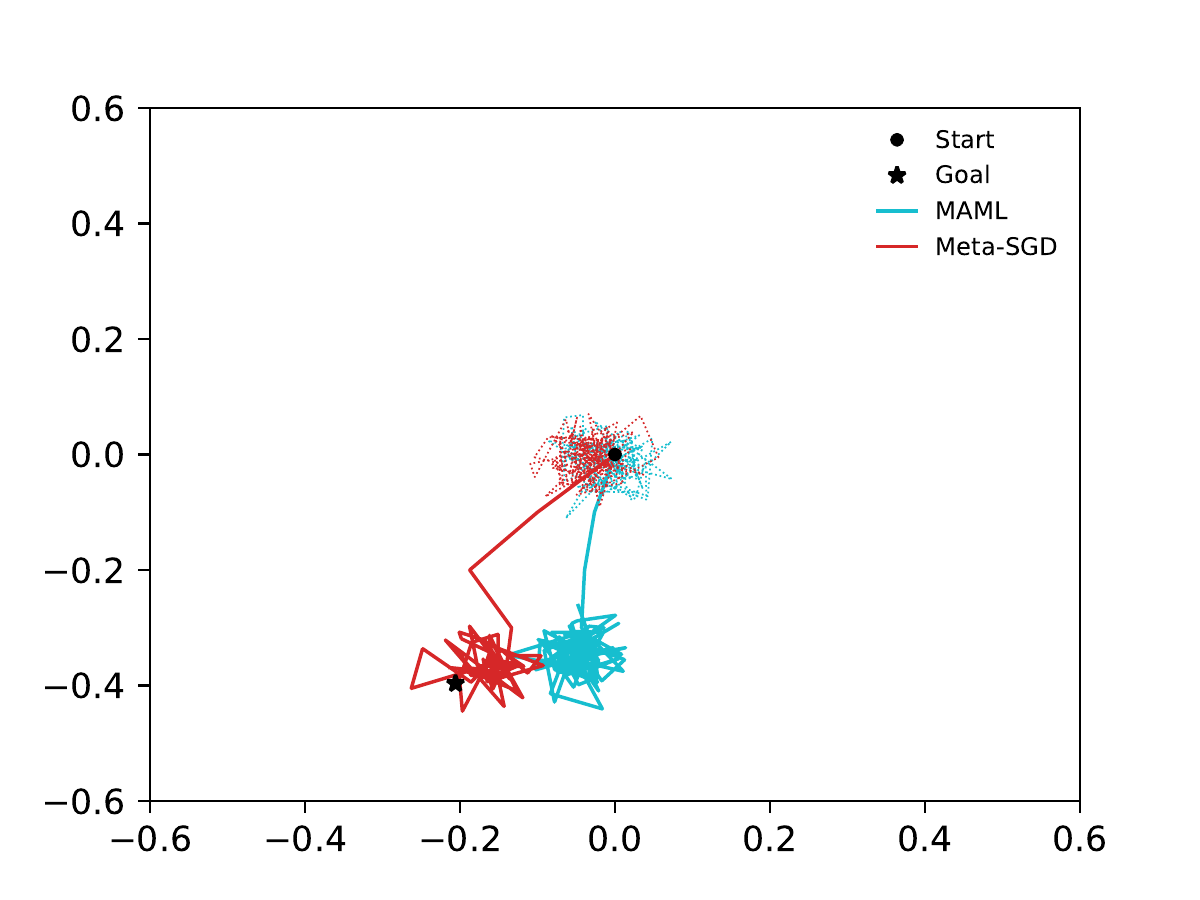}
	\end{subfigure}
	\begin{subfigure}{0.49\textwidth}
		\centering
		\includegraphics[width=1.0\textwidth]{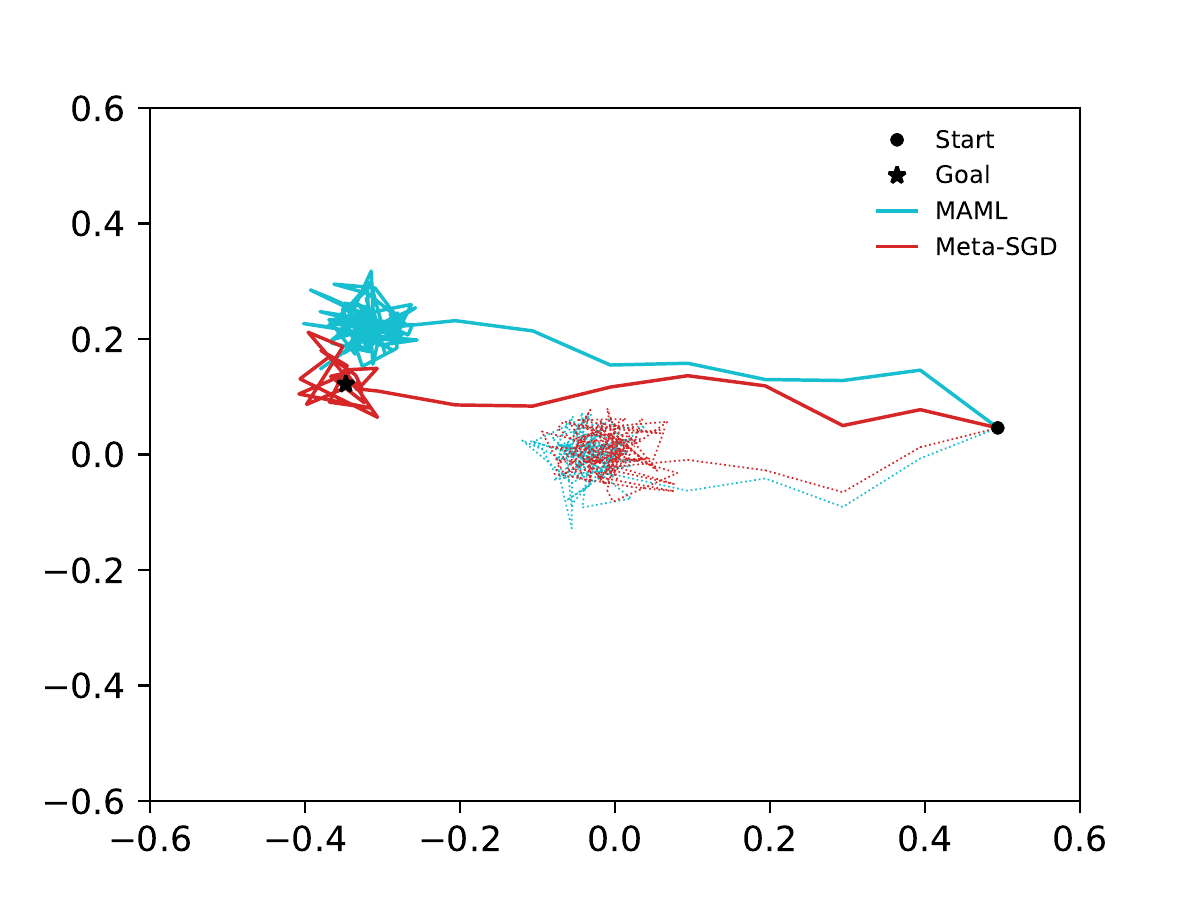}
	\end{subfigure}
	
	\caption{\textbf{Left:} Meta-SGD vs MAML on a 2D navigation task with fixed start position and randomly sampled goal position. \textbf{Right:} Meta-SGD vs MAML on a 2D navigation task with randomly sampled start and goal positions. }
	\label{fig:reinforcement_pictures}\vspace{-0.0cm}
\end{figure*}

In this experiment, we evaluate Meta-SGD on 2D navigation tasks, and compare it with MAML~\cite{finn2017model}. The purpose of this reinforcement learning experiment is to enable a point agent in 2D to quickly acquire a policy for the task where the agent should move from a start position to a goal position. We experiment with two sets of tasks separately. In the first set of tasks, proposed by MAML, we fix the start position, which is the origin $(0,0)$, and randomly choose a goal position from the unit square $[-0.5, 0.5]\times[-0.5, 0.5]$ for each task. In the second set of tasks, both of the start and goal positions are randomly chosen from the unit square $[-0.5, 0.5]\times[-0.5, 0.5]$.

Given a task, the state is the position of the agent in the 2D plane and the action is the velocity of the agent in the next step (unit time). The new state after the agent's taking the action is the sum of the previous state and the action. The action is sampled from a Gaussian distribution produced by a policy network, which takes the current state as input and outputs the mean and log variance of the Gaussian distribution. For the policy network, we follow \cite{finn2017model}. The mean vector is created from the state via a small neural network consisting of an input layer of size 2, followed by 2 hidden layers of size 100 with ReLU nonlinearities, and then an output layer of size 2. The log variance is a diagonal matrix with two trainable parameters. For the agent to move to the goal position, we define the reward as the negative distance between the state and the goal.

For meta-training, we sample 20 tasks as a mini-batch in each iteration. We first sample 20 trajectories per task according to the policy network and each trajectory terminates when the agent is within 0.01 of the goal or at the step of 100. Next, we use vanilla policy gradient~\cite{williams1992simple} to compute the empirical policy gradient $\nabla\mathcal{L}(\bm{\theta})$ and apply $\bm{\theta}^\prime = \bm{\theta} - \bm{\alpha}\circ\nabla\mathcal{L}(\bm{\theta})$ to update the policy network. After that, we sample 20 trajectories according to the updated policy network. Finally, we use Trust Region Policy Optimization~\cite{schulman2015trust} to update $\bm{\theta}$ and $\bm{\alpha}$ for all 20 tasks. For additional optimization tricks, we follow \cite{finn2017model}. We take 100 iterations in total.

For meta-testing, we randomly sample 600 tasks. For each task, we sample 20 trajectories according to the policy network initialized by the meta-learner, and then update the policy network by the vanilla policy gradient and the meta-learner. To evaluate the performance of the updated policy network on this task, 20 new trajectories are sampled and we calculate the return, the sum of the rewards, for each trajectory and take average over these returns as the average return for this task. The results averaged over the sampled 600 tasks with $95\%$ confidence intervals are summarized in Table~\ref{tab:res_reinforcement}, which show that Meta-SGD has relatively higher returns than MAML on both sets of tasks.

\begin{table}[h]
  \centering
  \begin{center}
    \caption{Meta-SGD vs MAML on 2D navigation}
    \label{tab:res_reinforcement}
    \vspace{0.1cm}
    \begin{tabular}{|c|c|c|}
      \hline
      & fixed start position & varying start position \\
      \hline
      MAML & $-9.12 \pm 0.66$ & $-10.71 \pm 0.76$ \\
      \hline
      Meta-SGD & $\mathbf{-8.64 \pm 0.68}$ & $\mathbf{-10.15 \pm 0.62}$ \\
      \hline
    \end{tabular}
  \end{center}
\vspace{-0.0cm}
\end{table}

Figure~\ref{fig:reinforcement_pictures} shows some qualitative results of Meta-SGD and MAML. For the set of tasks with fixed start position, the initialized policies with Meta-SGD and MAML perform quite similar -- the agents walk around near the start position. After one step update of the policies, both of the agents move to the goal position, and the agent guided by Meta-SGD has a stronger perception of the target. For the set of tasks with different start positions, the initialized policies with Meta-SGD and MAML both lead the agents to the origin. The updated policies confidently take the agents to the goal position, and still, the policy updated by Meta-SGD performs better. All these results show that our optimization strategy is better than gradient descent.



\section{Conclusions}

We have developed a new, easily trainable, SGD-like meta-learner Meta-SGD that can learn faster and more accurately than existing meta-learners for few-shot learning. We learn all ingredients of an optimizer, namely initialization, update direction, and learning rate, via meta-learning in an end-to-end manner, resulting in a meta-learner with a higher capacity compared to other optimizer-like meta-learners. Remarkably, in just one step adaptation, Meta-SGD leads to new state-of-the-art results on few-shot regression, classification, and reinforcement learning.  

One important future work is large-scale meta-learning. As training a meta-learner involves training a large number of learners, this entails a far more  computational demand than traditional learning approaches, especially if a big learner is necessary when the data for each task increases far beyond ``few shots''. Another important problem regards the versatility or generalization capacity of meta-learner, including dealing with unseen situations such as new problem setups or new task domains, or even multi-tasking meta-learners. We believe these problems are important to greatly expand the practical value of meta-learning.

{\small
\setlength{\bibsep}{5pt}
\bibliography{meta-learning}
\bibliographystyle{myplainnat}
}

\end{document}